\documentclass[letterpaper, 10 pt, conference]{ieeeconf}

\IEEEoverridecommandlockouts                             
\usepackage{cite}
\usepackage{epsfig} 
\usepackage{epstopdf}
\usepackage{amsmath}
\usepackage{bm}
\usepackage{soul}
\usepackage{amsfonts}
\usepackage{color}
\usepackage{listings}
\usepackage{subfigure}
\usepackage{graphicx}
\usepackage{wrapfig}
\usepackage{lscape}
\usepackage{rotating}
\usepackage{dirtytalk}
\usepackage{url}
\usepackage{hyperref}
\usepackage[makeroom]{cancel}
\usepackage{pifont}

\usepackage{graphicx}
\usepackage{algorithm}
\usepackage{algorithmic}

\usepackage{mdframed}

\title{
Agile Actions with a Centaur-Type Humanoid: A Decoupled Approach
}

\author{Matteo Parigi Polverini, Enrico Mingo Hoffman, Arturo Laurenzi, and  Nikos G. Tsagarakis%% <-this % stops a space
\thanks{This work was supported by the European  Union’s  Horizon  2020  Research  and Innovation  Program  under  Grant No. 779963 (EUROBENCH).}
\thanks{The authors are with the Humanoids \& Human Centered Mechatronics Research Line (HHCM), Istituto Italiano di Tecnologia (IIT), Genova, Italy {\tt\footnotesize \{matteo.parigi, enrico.mingo, arturo.laurenzi, nikos.tsagarakis\}@iit.it}}%
}

\begin{document}

\maketitle

%%%%%%%%%%%%%%%%%%%%%%%%%%%%%%%%%%%%%%%%%%%%%%%%%%%%%%%%%%%%%%%%%%%%%%%%%%%%%%%%
\begin{abstract}
The kinematic features of a centaur-type humanoid platform, combined with a powerful actuation, enable the experimentation of a variety of agile and dynamic motions. 
However, the higher number of degrees-of-freedom and the increased weight of the system, compared to the bipedal and quadrupedal counterparts, pose significant research challenges in terms of computational load and real implementation. 
To this end, this work presents a control architecture to perform agile actions, conceived for torque-controlled platforms, which decouples for computational purposes \mbox{offline} optimal control planning of lower-body primitives, based on a template kinematic model, and online control of the upper-body motion to maintain balance. Three stabilizing strategies are presented, whose performance is compared in two types of simulated jumps, while experimental validation is performed on a half-squat jump using the CENTAURO robot.
\end{abstract}

%%%%%%%%%%%%%%%%%%%%%%%%%%%%%%%%%%%%%%%%%%%%%%%%%%%%%%%%%%%%%%%%%%%%%%%%%%%%%%%%
\section{Introduction} \label{sec:introduction}
%%%%%%%%%%%%%%%%%%%%%%%%%%%%%%%%%%%%%
% MOVED HERE TO BE IN PAGE 2
\begin{figure*}
\vspace{3mm}
    \centering
    \includegraphics[width=.75\textwidth]{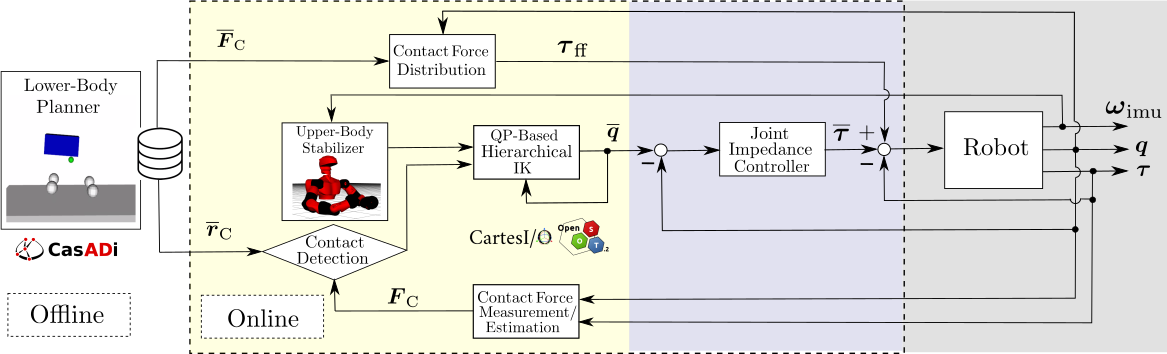}
     \caption{Overview of the proposed decoupled architecture. 
     }
     \label{fig:architecture}
\end{figure*}
%%%%%%%%%%%%%%%%%%%%%%%%%%%%%%%%%%%%%
The problem of realizing agile and dynamic actions 
% aiming at mimicking and possibly outperforming the biological example
with a legged robotic platform is an active research topic spanning: fast dynamic gaits \cite{nagasaka2004integrated, gehring2016practice}, jumping \cite{hutter2017anymal, katz2019mini}, kicking \cite{wensing2013generation}. 
In this respect, remarkable demonstrators have been realized thanks to the recent technological advances in designing powerful, robust and increasingly lightweight robotic platforms.
%, e.g. Boston Dynamics Atlas or the small-size quadruped MIT Mini Cheetah \cite{katz2019mini}. % and Solo \cite{grimminger2020open}.
%%%%%%%%%%%%%%%%%%%%%%%%%%%%%%%%%%%%%%%%%%%
%\par
The seminal work of Raibert on the one-legged hopper \cite{raibert1986legged}, followed by the jumping and landing pneumatic robot Mowgli \cite{niiyama2007mowgli}, represent the first examples of actuated jumping mechanisms. From then on, research on agile motions have been carried out, almost in parallel, on bipeds and quadrupeds.
%%%%%%%%%%%%%%%%%%%%%%%%%%%%%%%%%%%%%%%%%%%
%\par
Regarding the former type of platforms, 
%%%
Boston Dynamics has shown Atlas running outdoor and performing the world-famous back-flip and parkour demonstrators. 
%\footnote{\url{https://www.bostondynamics.com/atlas}}
Implementation details have not been made available, however in \cite{dai2014whole} an algorithm is presented which combines a simplified dynamic model with the robot full kinematics to make Atlas jump in simulation.
%%%
As shown in a popular video,
% \footnote{\url{https://www.youtube.com/watch?v=Bmglbk_Op64s}}
Honda Asimo has been capable of running and performing small jumps, even on single stance, 
%%%
while the small-size biped QRIO, manufactured by SONY, was able to run and jump through generation of dynamically consistent motion patterns \cite{nagasaka2004integrated}. 
%%%
In \cite{sakka2005humanoid} an  approach  based  on  ground reaction  forces is introduced to  make  the  \mbox{HRP-2}  robot  perform  vertical jumps  in  simulations.  
% Feet  reaction  forces remain constant during the entire jumping phase and directly depend on the desired flight height.
%%%
Recently, in \cite{bergonti2019torque} a torque and velocity controllers to perform jumps have been tested 
%both in simulation and 
on the humanoid robot iCub.
%%%%%%%%%%%%%%%%%%%%%%%%%%%%%%%%%%%%%%%%%%%
%\par
Almost concurrently, impressive results have been achieved among the quadrupedal robotic community.
%%%
In \cite{gehring2016practice} the ANYmal predecessor, the StarlETH quadruped using series elastic actuators, was able to perform dynamic gaits and a squat jump. Similar results have been demonstrated on ANYmal \cite{hutter2017anymal} by setting user-defined base target positions % (the \say{naive} approach) 
to the whole-body controller in order to perform the jump.
%%%
Thanks to recent advances in actuator development, pseudo-direct-drive concepts have found application in high-dynamic legged robots as in the  MIT Cheetah \cite{seok2013design,park2015online}, which is able to run and jump at high speeds. 
Similar concepts have been adopted in the small-size quadrupeds Minitaur  \cite{kenneally2016design} and MIT Mini Cheetah \cite{katz2019mini}. 
%The latter is able to perform back-flips using the open-source CasADi library \cite{Andersson2019} to set up the trajectory optimization problem for the back-flip motion.
%\footnote{\url{https://www.youtube.com/watch?v=xNeZWP5Mx9s}}. 

\section{Decoupled Architecture for Centaur-Type Humanoids} \label{sec:architecture}
%%%%%%%%%%
Centaur-type humanoids, as the wheeled-legged system CENTAURO \cite{kashiri2019centauro}, represent a new kinematic paradigm that combines the inherent stability advantage of quadrupeds over bipeds, with the loco-manipulation capabilities enabled by a humanoid upper-body. In this respect, while several agile actions are theoretically enabled by such topology (if combined with a powerful actuation), the increased weight of the system and the higher number of degrees-of-freedom (DoF), compared to the bipedal and quadrupedal counterparts, pose significant research challenges. 
Targeting the actual realization of dynamic behaviours on a centaur-type humanoid while meeting computational and implementation requirements, 
we present a control architecture, designed for torque-controlled platforms, which builds upon our previous work in \cite{Polverini2020}. 
It consists of the following components, organized in a \textit{decoupled} structure, see Fig. \ref{fig:architecture}:
\begin{itemize}
    \item[-] Offline stage: 
    a \textit{lower-body planner} employs optimal control, applied to a quadrupedal template kinematic model, to generate a database of lower-body agile primitives, consisting in a time series of planned contact positions  \mbox{$\overline{\bm r}_{\text C} \in \mathbb R^{k \times N_s}$} and contact forces \mbox{$\overline{\bm F}_{\text C}  \in \mathbb R^{k\times N_s}$}. 
    % Due to the wheeled-legged nature of CENTAURO, 
    We will hereafter assume  $n_\text{C}$  point contacts, thus \mbox{$k=3n_\text{C}$}, while \mbox{$N_s \in \mathbb{R}$} is the number of shooting intervals. % used in the OCPs transcription. 
    \item[-] Online stage: 
    the planned lower-body actions are replayed on the system using upper-body motions to maintain balance. 
    %This control layer further provides robustness w.r.t. the environment surface location and to external disturbances.
    The following components are employed:
    \begin{enumerate}
        \item \textit{Upper-body stabilizer}: is responsible to maintain the balance of the whole system through upper-body motions, based on IMU angular velocity \mbox{$\bm \omega_{\text{imu}} \in \mathbb{R}^3$} feedback;
        %Three possible feedback strategies are presented and compared;
        \item \textit{Hierarchical Inverse Kinematics (IK)}: 
        tracks the planned lower-body contact positions \mbox{$\overline{\bm r}_{\text C}$} and the upper-body stabilizing action by means of Quadratic Programming (QP);
        \item \textit{Contact force distribution}: 
        tracks at best the planned contact forces%\mbox{$\overline{\bm F}_{\text C}$}
        , while ensuring  balance. 
        \item \textit{Contact detection}: detects if contact with the environment has been established, based on estimated or measured contact forces \mbox{$\bm F_{\text C}\in \mathbb{R}^k$}.
        \item \textit{Joint-level control}: a decentralized joint impedance controller with torque feed-forward  \mbox{$\bm \tau_{\text{ff}} \in \mathbb R^n$}, feeding the torque control loop.
    \end{enumerate}
\end{itemize}
%%%%%%%%%%%%
While the lower-body planner and the upper-body stabilizer, which are main contributions of this paper, will be described in Sec. \ref{sec:lower_body_planner} and Sec. \ref{sec:upper_body_stabilizers}, respectively, 
the reader can refer to \cite{Polverini2020} for details on the remaining  components.

\section{Planning Lower-Body Actions} \label{sec:lower_body_planner}
%%%%%%%%%%%%%%%%%%%%%
\subsection{Lower-Body Template Kinematic Model}
\begin{figure}[!b]
    \centering
    \includegraphics[width=\columnwidth, trim={5cm 4.5cm 5cm 5cm}, clip]{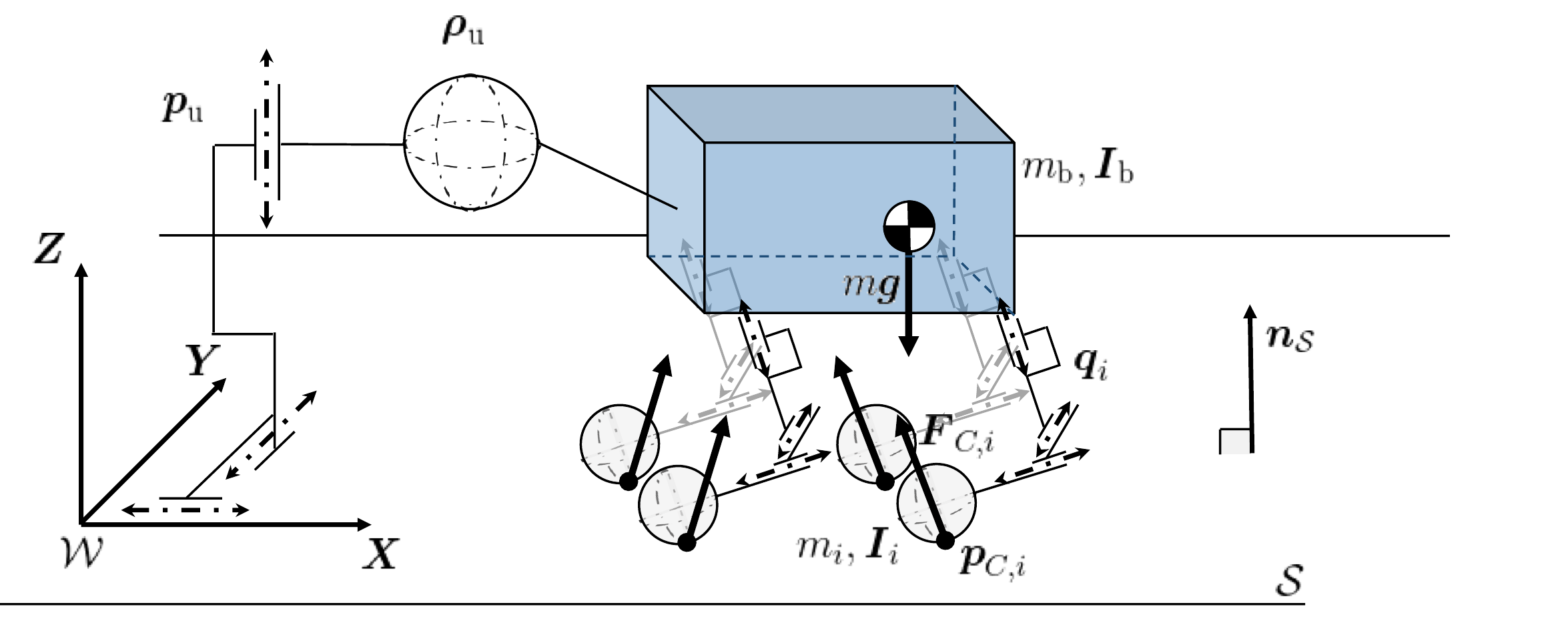}
    \caption{The 5-mass template model for the quadrupedal lower-body featuring prismatic joints.}
    \label{fig:template_lower_body}
\end{figure}

In order to overcome the computational complexity introduced by the adoption of a full kinematic model, the quadrupedal lower-body of a centaur-type platform will be herein modeled as a template $5$-mass floating-base system.
Note that, while this modelling choice simplifies the robot kinematics for computational purposes, no further simplification will be introduced on the dynamic model. 
With reference to Fig. \ref{fig:template_lower_body}, 
let us consider a floating-base system consisting of the actuated \textit{prismatic} joint coordinates 
\mbox{$\bm{q}_{\text{C}} \in \mathbb{R}^{k}$}, 
enabling the positioning of the $4$ leg end-effectors.
Again, as a common assumption for quadrupeds, point contacts are considered, thus \mbox{$k=12$}.
The corresponding Cartesian positions and contact forces for the $i$-th leg, expressed w.r.t. the world frame $\mathcal{W}$, are denoted with \mbox{$\bm{p}_{\text{C,i}}(\bm q) \in \mathbb{R}^3$} and \mbox{$\bm{F}_{\text{C,i}} \in \mathbb{R}^3$}, respectively.
The pose of the under-actuated floating-base is modeled through three prismatic joints \mbox{$\bm{p}_{\text{u}} \in \mathbb{R}^3$} and a spherical joint, whose orientation is given by the unit quaternion  $\boldsymbol{\rho}_{\text{u}} \in \mathbb{R}^4$. 
As shown in Fig.~\ref{fig:template_lower_body}, the inertial properties of each rigid body are given by the related mass value and inertia tensor. Note also that, in order account for the presence of the robot upper-body, as in CENTAURO, the waist body mass has been shifted towards the front legs.
Finally, let us consider a generic environment plane $\mathcal{S}$, whose orientation is given by its normal \mbox{$\bm{n}_\mathcal{S} \in \mathbb{R}^3$}. 
%%%%%%%%%%%%%%%%%%%%
%%%%%%%%%%%%%%%%%%%%
By now denoting with $n$ and $n_u$ the number of actuated and unactuated degrees-of-freedom (DoFs), respectively,  
the generalized coordinates can be collected in the vector \mbox{$\bm{q}\in \mathbb{R}^{n+n_u}$}, with $n=k$ and $n_u=7$ for the introduced template model:
\begin{equation}
    \bm{q} = 
    \begin{bmatrix}
        \bm{p}_{\text{u}}^T &
        \boldsymbol{\rho}_{\text{u}}^T & 
        \bm{q}_{\text{C}}^T 
    \end{bmatrix}^T
    \label{eq:joint_state}
\end{equation}
while the generalized velocities \mbox{$\bm{\nu}\in \mathbb{R}^{n+n_u-1}$} and accelerations \mbox{$\bm{\dot{\nu}}\in \mathbb{R}^{n+n_u-1}$} are given by:
\begin{subequations}
\begin{align}
    & \begin{matrix}
        \bm{\nu} = 
        \begin{bmatrix}
            \bm{\dot{p}}_{\text{u}}^T &
            \boldsymbol{\omega}_{\text{u}}^T & 
            \bm{\dot{q}}_{\text{C}}^T 
        \end{bmatrix}^T
         \end{matrix} \\
  & \begin{matrix}
        \bm{\dot{\nu}} = 
        \begin{bmatrix}
            \bm{\ddot{p}}_{\text{u}}^T & 
            \boldsymbol{\dot{\omega}}_{\text{u}}^T & 
            \bm{\ddot{q}}_{\text{C}}^T 
        \end{bmatrix}^T
    \end{matrix}
\end{align}
\end{subequations}
where \mbox{$\bm{\dot{p}}_{\text{u}}, \  \bm{\ddot{p}}_{\text{u}} \in \mathbb{R}^3$} and \mbox{$\boldsymbol{\omega}_{\text{u}}, \ \dot{\boldsymbol{\omega}}_{\text{u}} \in \mathbb{R}^3$} are the linear and angular velocity and acceleration, respectively, of the robot floating-base expressed in the world $\mathcal{W}$ coordinates.
%%%
Given \mbox{$\bm \rho_\text{u}=\begin{bmatrix} {\bm\epsilon}_{\text{u}}^T & {\eta}_{\text{u}}\end{bmatrix}^T$}, with \mbox{${\bm\epsilon}_{\text{u}}\in \mathbb R^3$} and \mbox{${\eta}_{\text{u}}\in \mathbb R$}, the quaternion propagation~\cite{graf2008} is given by:
\begin{equation}
    \bm{\dot{\rho}}_\text{u} = \left[ \frac{1}{2}\boldsymbol{\omega}_{\text{u}}, \ 0\right] \circ \bm{\rho}_\text{u}
    \label{eq:quaternion_propagation_global}
\end{equation}
expressing the relation between $\dot{\bm \rho}_\text{u}$ and $\bm\omega_{\text{u}}$.
The symbol $\circ$ is used to denote the quaternion product.
%%%%%%%%%%%%%%%%%%%%%%%%%%%%%%%%%
%%%%%%%%%%%%%%%%%%%%%%%%%%%%%%%%%
\subsection{Floating-Base Dynamic Model} \label{subsec:floating_base_dynamics}
The dynamics of a floating-base robot is expressed by the following equation:
\begin{equation}
    \label{eq:floatingdyn}
    \bm{B}(\bm{q})\bm{\dot{\nu}} + \bm{h}(\bm{q}, \bm{\nu}) = \bm{S}\bm{\tau} + \bm{J}_\text{C}^T(\bm{q})\bm{F}_\text{C} 
\end{equation}
where \mbox{$\boldsymbol{\tau} \in \mathbb{R}^{n}$} are the actuated joint torques,
while \mbox{$\bm{B}(\bm{q}) \in \mathbb{R}^{(n+n_u-1) \times (n+n_u-1)}$} is the full joint space inertia matrix and \mbox{$\bm{h}(\bm{q}, \bm{{\nu}}) \in \mathbb{R}^{n+n_u-1}$} is the vector of non-linear (gravity, centrifugal/Coriolis) terms.
%%%
Differently from a fixed-base robot, the actuation matrix \mbox{$\bm{S}\in\mathbb{R}^{(n+n_u-1)\times n}$} models the system under-actuation.
%%%
Contact forces \mbox{$\bm{F}_\text{C}\in\mathbb{R}^k$} are taken into account by concatenating the Jacobian of all support links \mbox{$\bm{J}_\text{C}(\bm{q})\in\mathbb{R}^{k\times (n+n_u-1)}$} and the corresponding overall contact wrench. 
The equation of motion in \eqref{eq:floatingdyn} can be further split into $n_u$ under-actuated and $n$ actuated rows, denoted with subscript $u$ and $a$, respectively.
\begin{subequations}
\begin{align}
    \bm{B}_u(\bm{q})\bm{\dot{\nu}} + \bm{h}_u(\bm{q}, \bm{\nu}) &=  \bm{J}_\text{C,u}^T(\bm{q})\bm{F}_\text{C} \label{eq:floatingdyn_unact}\\
    \bm{B}_a(\bm{q})\bm{\dot{\nu}} + \bm{h}_a(\bm{q}, \bm{\nu}) &= \bm{\tau} + \bm{J}_\text{C,a}^T(\bm{q})\bm{F}_\text{C}  \label{eq:floatingdyn_act}
\end{align}
\end{subequations}
%%%%%%%%%%%%%%%%%%%%%%%%%%%%%%%%%
%%%%%%%%%%%%%%%%%%%%%%%%%%%%%%%%%
\subsection{Optimal Control Formulation and Transcription}
Let us consider the following choice for the state $\bm{x}(t)$ and control $\bm{u}(t)$ vectors:
\begin{subequations}
\begin{align}
    & \begin{matrix}
        \bm{x}  = 
        \begin{bmatrix}
            \bm{q}^T & 
             \bm{\nu}^T
        \end{bmatrix}^T 
    \end{matrix} 
    \\
    & \begin{matrix}
        \bm{u}  = 
        \begin{bmatrix}
            \bm{\dot{\nu}}^T&
            \bm{F}_{\text{C}}^T 
        \end{bmatrix}^T 
    \end{matrix}
\end{align}
\end{subequations}
%hence:
%\begin{equation}
%\bm{\dot{x}}  = 
%        \begin{bmatrix}
%            \bm{\dot{q}}^T & 
%             \bm{\dot{\nu}}^T
%        \end{bmatrix}^T 
%\end{equation}
%%%
Based on the floating-base model described in Sec. \ref{subsec:floating_base_dynamics}, the OCP formulation we address in this Section reads as:
\begin{mdframed}
%%%%%%%%%%%%%%%%%%%%%%%%%%%%%%%%%%%%%%
\begin{equation} 
\label{eq:nlp_Diehl}
\begin{aligned}
%%%% 
& \underset{\bm x(\cdot), \bm u(\cdot)}{\operatorname{min}} \ \ \int_0^T L\big(\bm x(t),\bm u(t)\big)dt + E\big(\bm x(T)\big)\\
& \text{subject to}\\ 
\end{aligned}
\end{equation}
\vspace{-0.5cm}
\begin{IEEEeqnarray*}{lc}
    \quad \bm{x}_{0} - \bm{x}^{\text{init}} = \bm{0} & \text{initial state}\\
    \quad \bm{x}({T}) - \bm{x}^{\text{goal}} = \bm{0} & \text{final state}\\
    % \vspace{-0.2cm}\\
    % %%%% 
    % \cline{1-2}\vspace{-0.5cm}\\
    %%%
    \quad \bm{\dot{x}}(t) - \bm f\big( \bm x(t),\bm u(t) \big) = \bm 0 \quad & 
    \text{double integrator}\\
    \quad \bm{\tau}_{\text{u}}\big( \bm x(t),\bm u(t) \big) = \bm 0 \quad & 
    \text{under-actuation}\\
    \quad \underline{\bm \tau} \le \bm{\tau}\big( \bm x(t),\bm u(t) \big) \le \overline{\bm \tau} \quad & 
    \text{torque bounds}\\
    % \vspace{-0.2cm}\\ 
    % %%%% 
    % \cline{1-2}\vspace{-0.5cm}\\
    %%%
    \quad \underline{\bm{{p}}}_\text{Ci} \le \bm{{p}}_\text{Ci}(t) \le \overline{\bm{{p}}}_\text{Ci} \quad & \text{work-space bounds}\\
    \quad \text{if Ci in contact:} \\
    \quad \quad \quad \bm{{p}}_\text{Ci}(t \in T_{\mathcal{C}}) \in \mathcal{S}(\bm{{p}}_\text{Ci}) \quad & \text{surface contact}\\
    \quad \quad \quad \dot{\bm{{p}}}_\text{Ci}(t \in T_{\mathcal{C}}) = \bm{0} \quad &  \text{no slip condition}\\
    \quad \quad \quad \bm{F}_{\text{Ci}}(t \in T_{\mathcal{C}} )  \in \mathcal F(\bm{F}_{\text{Ci}}, \bm n_\mathcal{S}, \mu) \quad & \text{friction cone} \\
    \quad \text{otherwise:} \\
    \quad \quad \quad \bm{F}_{\text{Ci}}(t \not\in T_{\mathcal{C}}) = \bm 0 \quad & \text{no contact force}
    % \vspace{-0.2cm}\\
    % \cline{1-2}\vspace{-0.5cm}
    %%%%
\end{IEEEeqnarray*}
\end{mdframed}
%%%
Herein the double-integrator relation from \mbox{$\dot{\bm \nu}(t)$} to \mbox{$\bm{q}(t)$} can be expressed through the following state-space representation:
\begin{equation}\label{eq:double_integrator}
    \bm{\dot{x}}(t) = \begin{bmatrix} \bm 0 & 
    \bm{\hat{S}}\\ 
    \bm 0 & \bm0\end{bmatrix} \bm{x}(t) 
    + \bm f_\text{quat}\big( \bm x(t)\big) + 
    \begin{bmatrix} \bm 0 & \bm 0  \\ \bm I & \bm 0 \end{bmatrix} \bm{u}(t)
\end{equation}
where $\bm f_\text{quat}(\bm x)$ represents the quaternion propagation in \eqref{eq:quaternion_propagation_global},
while the selection matrix $\bm{\hat{S}}$ is given by: 
\begin{equation}
\bm{\hat{S}} = \begin{bmatrix} \bm I_{3\times (n+n_u)} & \bm 0_{4\times (n+n_u)} & \bm I_{n \times (n+n_u)} \end{bmatrix}^T
\end{equation}
%%%%%%%%%%%
% \par
When the robot is establishing contact with a surface $\mathcal{S}$, constraints on contact positions, velocities and forces must be simultaneously enforced. 
To this end, the following constraint:
\begin{equation}
    \bm{{p}}_\text{Ci}(\bm q) \in \mathcal{S}\big(\bm{{p}}_\text{Ci}(\bm q)\big)
    \label{eq:surface_constraint}
\end{equation}
ensures that the $i$-th contact point lays on the surface $\mathcal{S}$.
In the simple planar case, the surface equation is given by:
\begin{equation}\label{eq:climbing_surface}
%\mathcal{S}\big(\bm{{p}}_\text{Ci}(\bm q)\big) : a\ p_{\text{Ci}}^{[x]} + b\ p_{\text{Ci}_y}^{[y]} + c\ p_{\text{Ci}}^{[z]} +d = 0
\mathcal{S}\big(\bm{{p}}_\text{Ci}(\bm q)\big) : {\bm n_{\mathcal{S}}}^T \bm{{p}}_\text{Ci} + d = 0
\end{equation}
being \mbox{$\bm n_{\mathcal{S}}=\begin{bmatrix} a & b & c \end{bmatrix}^T \in \mathbb{R}^3$} the surface normal, with $a,b,c,d \in \mathbb{R}$.
We also need to ensure that contact points do not slip, i.e.: 
\begin{equation}
    \bm{\dot{p}}_\text{Ci}(\bm q, \bm{\nu}) = \bm{J}_{\text{Ci}}\bm{\nu} = \bm{0}
    \label{eq:no_slip}
\end{equation}
%%%
In order to further encode the impact of the surface orientation on contact forces, friction constraints must be incorporated.
Being \mbox{$\bm{F}_\text{Ci}^n  \in \mathbb{R}^3$} and \mbox{$\bm{F}_\text{Ci}^t \in \mathbb{R}^3$}, the normal and tangential component of the contact force at the $i$-th contact point: 
\begin{subequations}
\begin{align}
    \bm{F}_\text{Ci}^n &= (\bm{F}_\text{Ci} \cdot \bm n_\mathcal{S})\bm n_\mathcal{S}\\
    \bm{F}_\text{Ci}^t &= \bm{F}_\text{Ci} - (\bm{F}_\text{Ci} \cdot \bm n_\mathcal{S})\bm n_\mathcal{S} 
\end{align}
\end{subequations}
the $i$-th point contact remains in rest contact mode if  $\bm{F}_\text{Ci}$ lies inside the friction cone directed by $\bm n_\mathcal{S}$, i.e.:
\begin{equation}\label{eq:friction_conic}
\mathcal F(\bm{F}_{\text{C}i}, \bm n_\mathcal{S}, \mu):=
\begin{cases}
\bm{F}_\text{Ci} \cdot \bm n_\mathcal{S}  > F_{\text{thr}}\\
\Arrowvert \bm{F}_\text{Ci}^t \Arrowvert_2  \leq \mu (\bm{F}_\text{Ci} \cdot \bm n_\mathcal{S})
\end{cases}
\end{equation}
where $\mu$ is the Coulomb friction coefficient, while \mbox{$F_{\text{thr}} \ge 0$} is a scalar force threshold. The Euclidean norm \mbox{$\Arrowvert \cdot \Arrowvert_2$} models Coulomb friction cones with circular section. 
%%%%%%%%%%%%%%%%%%%%%
% \par
The under-actuation constraint is implemented according to \eqref{eq:floatingdyn_unact} by denoting with \mbox{$\bm{\tau}_{\text{u}} \in \mathbb R^{n_u-1}$} the under-actuated torques, i.e.:
\begin{equation}
    \bm{\tau}_{\text{u}} = \bm{B}_u\bm{\dot{\bm{\nu}}} + \bm{h}_u -  \bm{J}_\text{C,u}^T\bm{F}_\text{C} 
\end{equation}
Similarly, torque bounds are imposed on the actuated torques \mbox{$\bm{\tau} \in \mathbb R^{n}$}, whose expression can been retrieved from \eqref{eq:floatingdyn_act}:
\begin{equation}
\bm \tau = \bm{B}_a\bm{\dot{\bm{\nu}}} + \bm{h}_a - \bm{J}_\text{C,a}^T\bm{F}_\text{C}
\end{equation}

%%%%%%%%%%%%%%%%%%%%
\begin{figure*}
    \centering
    \subfigure[Squat jump]
    {\includegraphics[width=\columnwidth, trim={0cm 0cm 0cm 0cm}, clip]{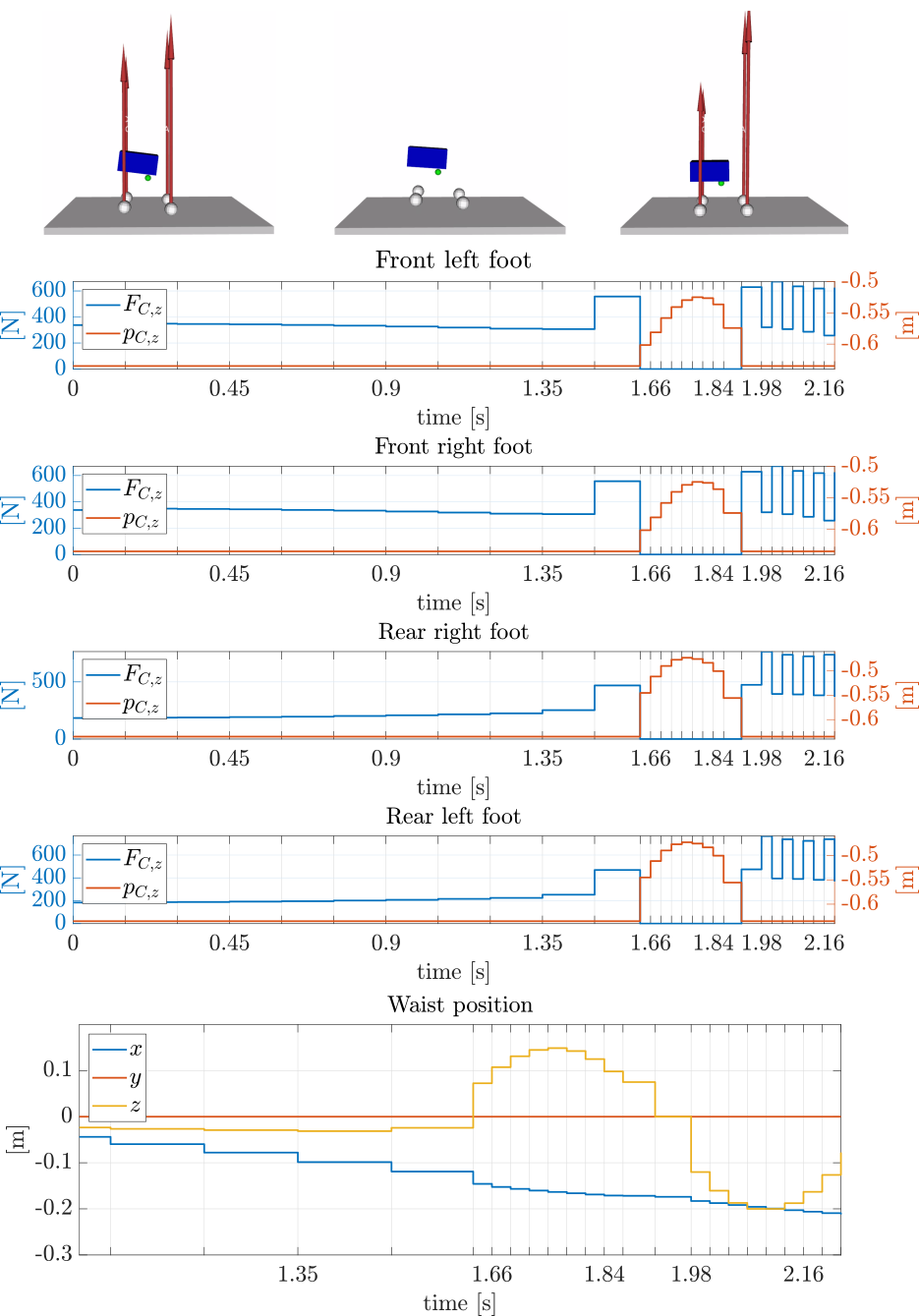} \label{fig:jump_plan_plot}}
    \subfigure[Half-squat jump]
    {\includegraphics[width=\columnwidth, trim={0cm 0cm 0cm 0cm}, clip]{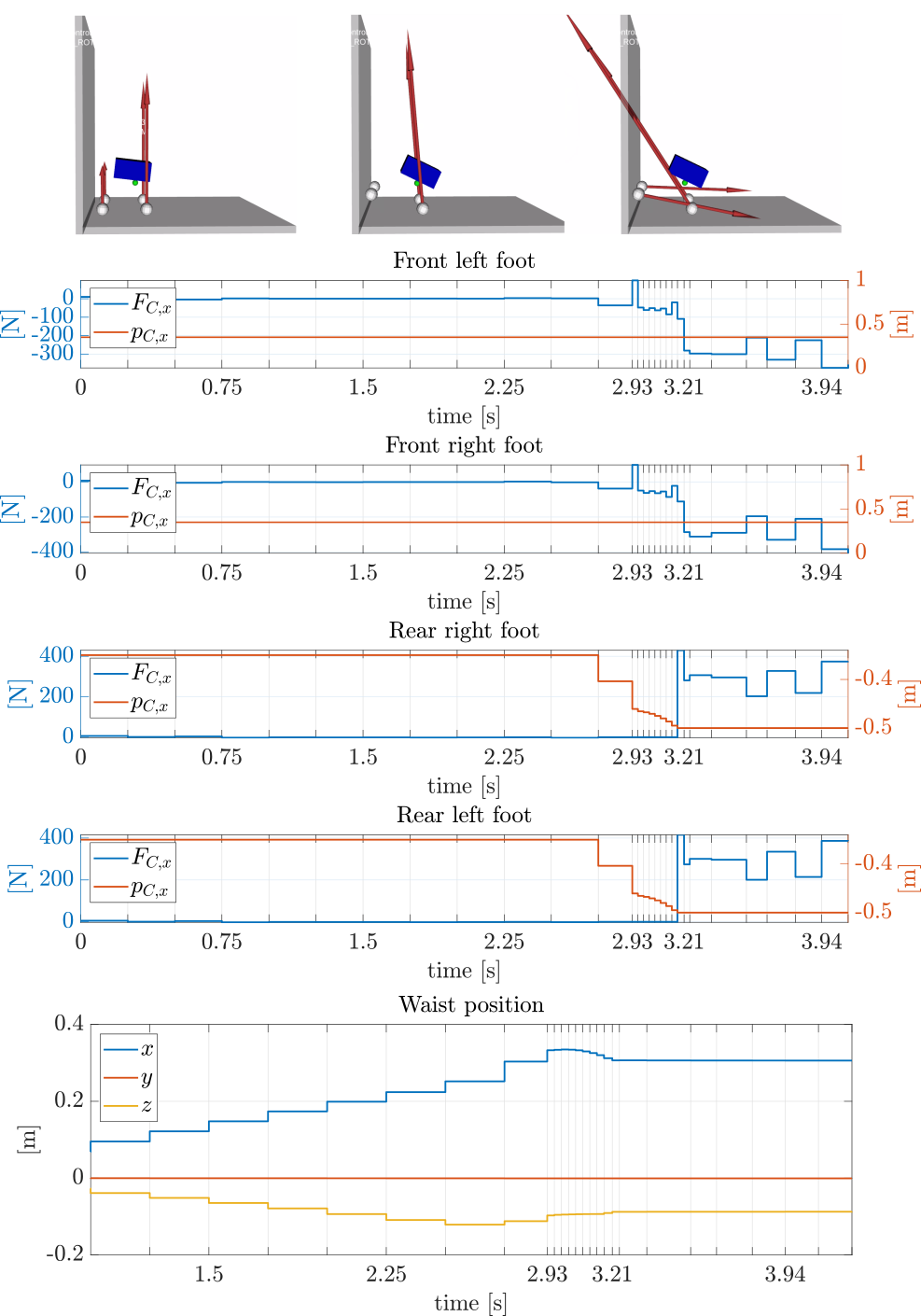} \label{fig:wall_plan_plot}}
    \caption{Snapshots of the produced motion in rviz ROS with visualization of contact forces (upper snapshots) and time histories of contact forces, feet positions and waist position (lower plots) for the squat jump (left side) and the half-squat jump (right side) planned agile actions.
    % Note how the step-size is adjusted by the solver. 
    }
    \label{fig:comparison}
\end{figure*}
%%%%%%%%%%%%%%%%%%%%%%%%%%%
\par
In order to perform a Direct Multiple Shooting (DMS) transcription \cite{diehl2006fast} of \eqref{eq:nlp_Diehl} into a Nonlinear Program (NLP) that can be solved by off-the-shelf solvers \cite{wachter2006implementation}, let us consider a number of shooting intervals $N_s$, which discretize the control horizon.
The state variable and control vector at the \mbox{$k$-th} shooting interval, $\bm{x}_\text{k}$ and $\bm{u}_\text{k}$ respectively, are denoted as:
\begin{subequations}
\begin{align}
    & \begin{matrix}
        \bm{x}_{\text{k}}  = 
        \begin{bmatrix}
            \bm{q_{\text{k}}}^T & 
            \bm{\nu_{\text{k}}}^T 
        \end{bmatrix}^T
    \end{matrix} 
    \\
    & \begin{matrix}
        \bm{u}_{\text{k}}  = 
        \begin{bmatrix}
            \bm{\dot{\nu}}_{\text{k}}^T&
            \bm{F}_{\text{C},\text{k}}^T 
        \end{bmatrix}^T
    \end{matrix}
\end{align}
\end{subequations}
We hereafter assume a piece-wise constant control parametrization along each shooting interval.
The $N_s+1$ states are collected in the state vector $\bm{X}$: 
\begin{equation}
\mbox{$\bm{X} = \left[\bm{x}_0^T, \bm{x}_1^T, \dots, \bm{x}_{N_s}^T\right]^T$}
\end{equation}
and the $N_s$ controls in the control vector \mbox{$\bm{U}$}:
\begin{equation}
\mbox{$\bm{U} = \left[\bm{u}_0^T, \bm{u}_1^T, \dots, \bm{u}_{N_s-1}^T\right]^T$}.
\label{eq:classic_implementation}
\end{equation}
%%%%%%
In agreement with DMS, a \textit{continuity condition} needs to be further satisfied:
\mbox{$\bm s\left( \bm{x}_{\text{k}},\bm{u}_{\text{k}} \right) - \bm{x}_{\text{k+1}} = \bm{0}$}.
Here the function \mbox{$ \bm s\left( \bm{x}_{\text{k}},\bm{u}_{\text{k}} \right)$} is used to simulate the double integrator dynamics in \eqref{eq:double_integrator} over one shooting interval.
%, starting from the state $\bm{x}_{\text{k}}$ and using the control input $\bm{u}_{\text{k}}$, and to match the result with the state variable at the following shooting interval $\bm{x}_{\text{k+1}}$.
%%%%%%
We additionally leave the solver free to decide the optimal step size for each shooting interval, introducing the step size variable $dt_{\text{k}} \in \mathbb{R}^+$ in the control vector:
\begin{equation}
    \bm{u}_{\text{k}}  = 
        \begin{bmatrix}
            \bm{\dot{\nu}}_{\text{k}}^T&
            \bm{F}_{\text{C},\text{k}}^T &
            dt_{\text{k}}
        \end{bmatrix}^T,
    \label{eq:variable_step_implementation}
\end{equation}
together with the bound:
\mbox{$\underline{dt} \leq dt_{\text{k}} \leq \overline{dt}$}.
%%%%%%%
Finally, we adopted a linearized version of friction cones which, in our experience, is easier to handle for the solver. 

\subsection{Planned Agile Actions}
Building upon the OCP in \eqref{eq:nlp_Diehl}, we hereafter address the planning problem of a set of agile lower-body actions, consisting in two types of squat jumps.
% \textcolor{red}{Note that, due to the wheeled-legged nature of the CENTAURO platform, fast gaits can be performed through wheeled locomotion.} 
The related NLPs have been implemented using CasADi~\cite{Andersson2019} and Pinocchio~\cite{pinocchioweb}%\footnote{The development on CasADi and Pinocchio that led to the realization of the present work has been collected into two software packages: the \emph{casadi\_kin\_dyn} library (\url{https://github.com/ADVRHumanoids/casadi_kin_dyn}) and the \emph{Horizon} library (\url{https://github.com/ADVRHumanoids/Horizon})}.
\footnote{Two software packages have been developed to this end: the \emph{casadi\_kin\_dyn} library (\url{https://github.com/ADVRHumanoids/casadi_kin_dyn}) and the \emph{Horizon} library (\url{https://github.com/ADVRHumanoids/Horizon})}.
The planned trajectories are then conveniently interpolated using the integrator function \mbox{$\bm s\left(\cdot\right)$} with fixed step-size of \mbox{$1\cdot 10^{-3}s$} and replayed on the robot in the online stage, see Fig. \ref{fig:architecture}.
%%%%%%%%%%
\subsubsection{Squat jump}
The robot is initially in contact with the ground, while its goal is to reach a target floating-base position \mbox{$\overline{\bm p}_{\text{u}}\in \mathbb R^3$} displaced vertically along the $z-$axis. Since the control horizon comprises also the landing phase, we denote $N_\text{T}$ as the take-off interval and $N_\text{L}$ as the landing interval, with: \mbox{$N_s = 30$}, \mbox{$N_\text{T} = 10$} and \mbox{$N_\text{L} = 20$}. In order to give the possibility to the solver to determine the duration of each phase, the control vector comprises the step-size variable % $dt_{\text{k}} \in \mathbb{R}^+$ 
within the following bounds: $\underline{dt}_\text{k} = 0.03 s$ and $\overline{dt}_\text{k} = 0.25 s$. 
%%%
The following piece-wise cost function has been considered:
\begin{align}\label{eq:cost_function_jump}
&F(\bm{X}, \bm{U}) =  
%%%
\sum_{\text{k}=N_\text{T}+1}^{N_\text{L}} \Big( \gamma_{\nu}\bm{\nu}_\text{k}^T\bm{\nu}_\text{k} + \gamma_{p_{\text{u}}}(p_{\text{u,k}}^{[\text{z}]}- \overline p_{\text{u}}^{[\text{z}]})^2 \Big) \ +\nonumber \\
%%%
&+\sum_{\text{k}=N_\text{T}+1}^{N_s} \gamma_{{\dot q}_{\text{C}}}\bm{\dot q}_{\text{C,k}}^T\bm{\dot q}_{\text{C,k}} +
\sum_{\text{k}=N_\text{L}+1}^{N_s-2}\gamma_{\dot{F}_\text{C}}\dot{\bm{F}}_\text{C,k}^T\dot{\bm{F}}_\text{C,k}
+ \sum_{\text{k}=0}^{N_s-2} \gamma_{\ddot{\nu}}\bm{\ddot{\nu}}_\text{k}^T\bm{\ddot{\nu}}_\text{k} 
\end{align}
with \mbox{$\gamma_{p_{\text{fb}}} = 50$}, \mbox{$\gamma_{{\dot q}_{\text{a}}} = 1$}, \mbox{$\gamma_{\dot{F}_\text{C}} = 1$} and \mbox{$\gamma_{\ddot{\nu}} = 10^{-3}$}.
%%%
We first penalize the tracking of a target floating-base position along the $z-$axis during the flight phase only, i.e. between the take-off and the landing shooting interval. 
The second term in~\eqref{eq:cost_function_jump} minimizes the velocity of the actuated joints during the flight phase, while the third and last terms minimize the derivative of contact forces (\mbox{$\dot{\bm{F}}_\text{C,k} \approx {\bm{F}}_\text{C,k+1}-{\bm{F}}_\text{C,k}$}) and of accelerations (\mbox{$\bm{\ddot{\nu}}_\text{k} \approx \bm{\dot{\nu}}_\text{k+1}-\bm{\dot{\nu}}_\text{k}$}), respectively, in order to smooth out the control actions.  
Snapshots of the resulting motion and time histories of relevant quantities are depicted in Fig. \ref{fig:jump_plan_plot} (these results are also illustrated in the accompanying video).
Due to the introduced offset on the waist mass location (modeling the presence of the upper-body), the robot waist moves backward before jumping, so that contact forces are equally distributed. Note also how the solver adapts the step-size.
%%%%%%%%%%%%%%%%%%%%

\subsubsection{Half-squat jump} % wall walks - rearing - prancing - wall kick jump
Inspired by the pushing task in \cite{Polverini2020}, where CENTAURO contacts a wall with its rear feet in order to push a heavy object, we set up an OCP in which the robot, initially in contact with the ground $\mathcal{S}_1$, has to  simultaneously lift and put its rear feet on a vertical surface $\mathcal{S}_2$ pivoting on its front feet, in a so-called \say{half-squat jump} manoeuvre.
The following parameter have been selected: \mbox{$N_s = 30$}, \mbox{$N_\text{T} = 10$} and \mbox{$N_\text{L} = 20$}, \mbox{$\underline{dt}_\text{k} = 0.03 s$}, \mbox{$\overline{dt}_\text{k} = 0.25 s$}. The cost function in \eqref{eq:cost_function_jump} has been employed, with same weights, now penalizing the tracking of the initial floating-base pose.
%%%%
Snapshots of the resulting motion and time histories of relevant quantities are depicted in Fig. \ref{fig:wall_plan_plot}. 
The robot waist moves towards the front feet in order to be able to lift the rear legs and consequently establish contact with the vertical surface.

\section{Upper-Body Stabilizers} \label{sec:upper_body_stabilizers}
We here present three different feedback strategies that exploit upper-body motion to maintain balance.
\begin{figure*}
    \centering
    \subfigure[Squat jump]
    {\includegraphics[width=.9\columnwidth, trim={0cm 0cm 0cm 0cm}, clip]{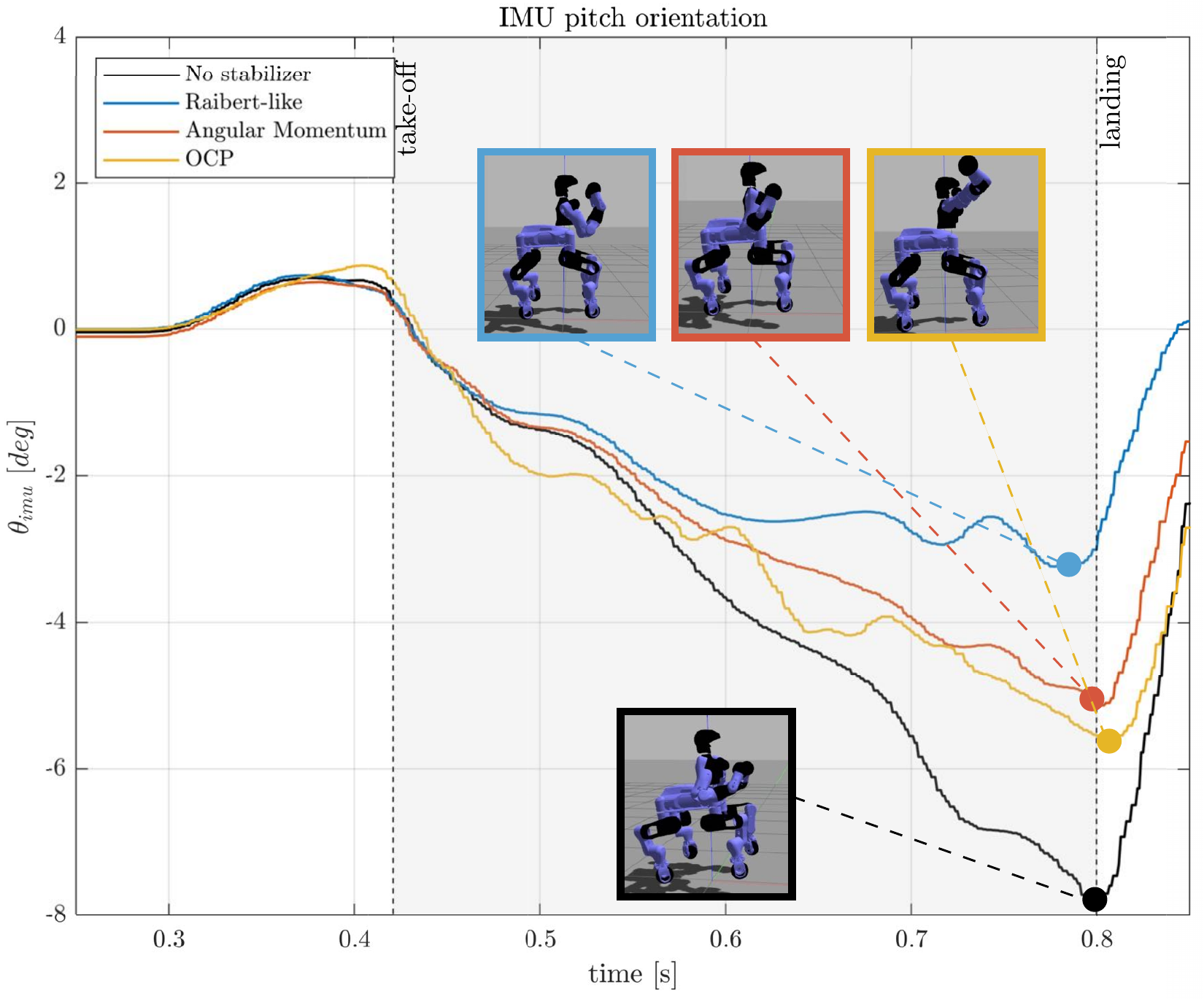} \label{fig:jump_comparison}}
    \subfigure[Half-squat jump]
    {\includegraphics[width=.9\columnwidth]{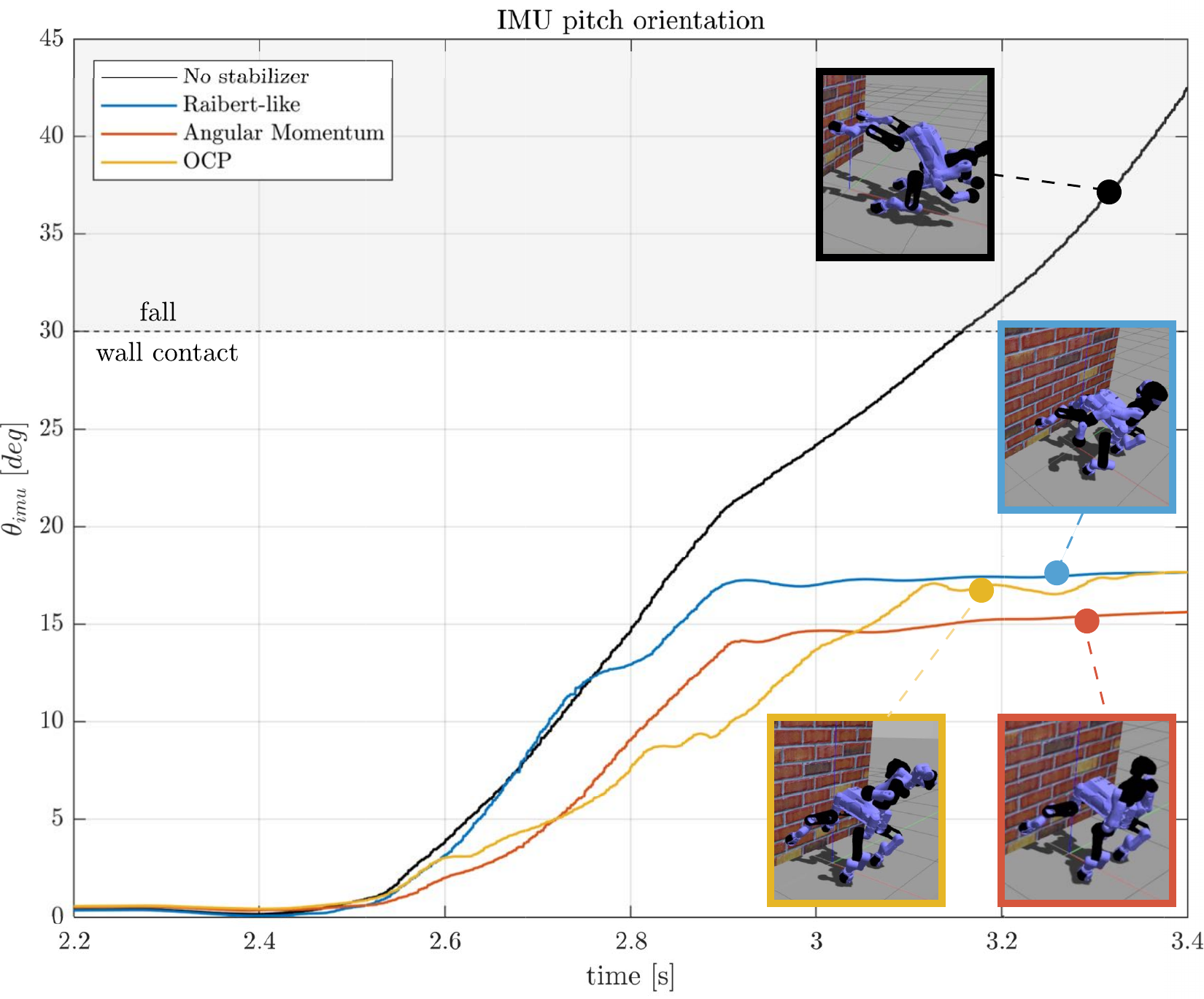} \label{fig:wall_comparison}}
    \caption{Time histories of the IMU pitch orientation and snapshots from Gazebo simulations of the squat jump (left side) and the half-squat jump (right side) planned agile actions, with and without (black solid lines) the proposed upper-body stabilizing strategies.}
    \label{fig:comparison_sim}
\end{figure*}
%%%%%%%%%%%%%%%%%%%%
\begin{figure*}
    \centering
    \subfigure[No stabilizer]
    {\includegraphics[width=\columnwidth, trim={0cm 0cm 0cm 0cm}, clip]{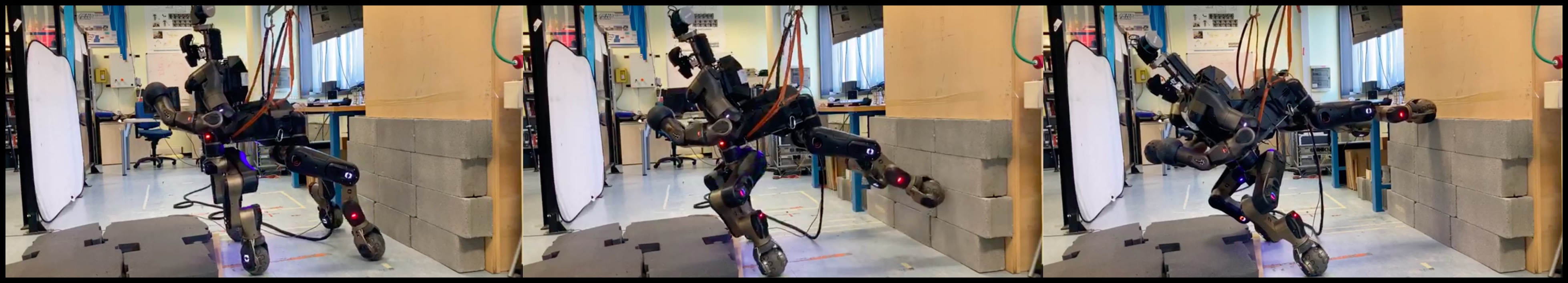} \label{fig:no_stab_exp}}
    \subfigure[Raibert-like stabilizer]
    {\includegraphics[width=\columnwidth, trim={0cm 0cm 0cm 0cm}, clip]{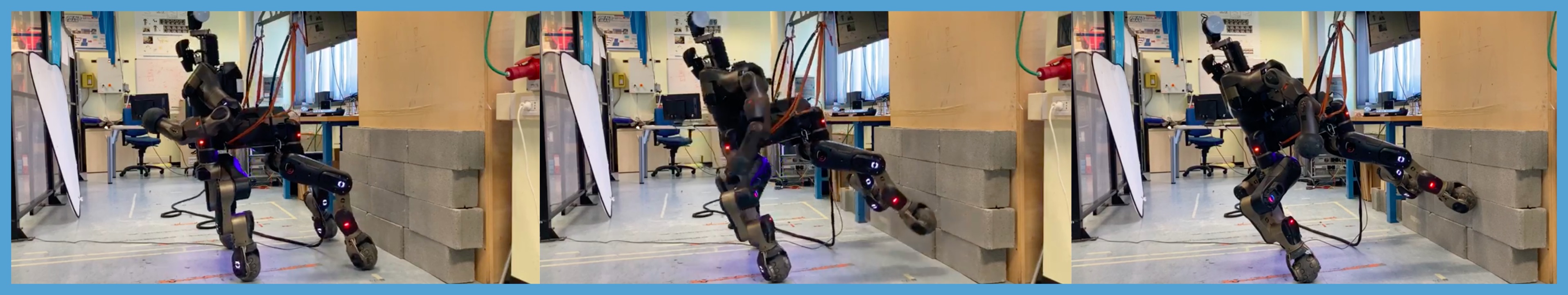} \label{fig:Raibert_exp}}
    \caption{Snapshots from performed half-squat jumps experiments with no stabilizing action (left side) and employin with the proposed Raibert-like postural task (right side). The arms' motion effectively prevents the robot from tilting excessively, thus allowing to establish a steady contact with the wall.}
    \label{fig:comparison_exp}
\end{figure*}
%%%%%%%%%%%%%%%%%%%%%%%%%%%%%%%
%The kinematics simplification introduced by the lower-body planner, comes at the cost of neglecting the impact of the upper-body configuration during the replay of the planned trajectories. As it will be shown in Sec. \ref{sec:simulations} for the case of CENTAURO, this results in a rotation of the whole system, predominantly about the $y$-axis, during flight phases. In order to mitigate this effect, we here present three different feedback strategies that exploit upper-body motion to maintain balance.
%, based on IMU feedback.
% As it will be shown in Sec. \ref{sec:simulations}, due to the presence of a humanoid upper-body, like in CENTAURO, the system is subject to a whole-body rotation, predominantly about the $y$-axis, during the flight phase of the planned agile actions. Based on this consideration, we here present three different stabilizing strategies that exploit upper-body motion to maintain balance, based on IMU feedback.

\subsection{Raibert-Like Postural Task}\label{subsec:raibert-like}
In the work of Raibert on the one-legged hopper \cite{raibert1986legged}, body posture control on ground is obtained by setting the torque between the leg and the body to be proportional to the body angle. As a result, the balance arm, acting as fly wheel, spins in the direction of falling in order to stabilize the system posture.
% http://www.cs.cmu.edu/~cga/humanoids-ugrad/
% http://www.animats.com/papers/leggedrun/leggedrun.html
%%%%%%%%%%%
%\par
Applying a similar control approach to the upper-body of a humanoid, the following postural task can be introduced:
%%%%%%%%%%%%%%%%%%%%%%%%%%%%%%%%
\begin{equation}\label{eq:posture_raibert}
\mathcal{T}_{\substack{\text{Posture-Raibert}}}: 
%\bm{S}_{\text{ub}} \bm q_{\text{d,k+1}} = \bm{S}_{\text{ub}} \bm q_{\text{d,k}} + \bm{\Gamma} \ \omega_{\text{imu}}^{[P]}
\bm{S}_{\text{ub}} \bm q_{\text{d,k+1}} = \bm{S}_{\text{ub}} \bm q_{\text{d,k}} + 
\bm{\Gamma} \bm \omega_{\text{imu}}
\end{equation}
%%%%%%%%%%%%%%%%%%%%%%%%%%%%%%%%
where \mbox{$\bm q_{\text{d}}\in \mathbb{R}^n$} is the desired robot posture (initialized in the homing configuration), the matrix \mbox{$\bm{S}_{\text{ub}} \in \mathbb{R}^{n\times n}$} enables the motion of a subset of the upper-body joints, 
%\mbox{$\omega_{\text{imu}}^{[P]} \in \mathbb{R}$} 
\mbox{$\bm \omega_{\text{imu}} \in \mathbb{R}^3$} 
is the measured IMU angular velocity, while \mbox{$\bm{\Gamma} \in \mathbb{R}^{n\times 3}$} is a gain matrix.
% \par
The considered SoT can be written using the Math of Task (MoT) formalism \cite{OpenSot17} as follows:
\begin{equation}
\small
\begin{pmatrix}
\left(\sum_i \   ^{\text{Pelvis}}\mathcal{T}_{{\text{Foot}}_i}^\text{[XYZ]}\right)/\\
%\\
 \mathcal{T}_{\substack{\text{Posture-Raibert}}} \\ 
\end{pmatrix}
<<
\begin{pmatrix}
\mathcal{C}_{\substack{\text{Pos.}\\\text{Lims}}}+\mathcal{C}_{\substack{\text{Vel.}\\\text{Lims}}}
\end{pmatrix},
\label{eq:box_stack_Raibert}
\end{equation}
%%%%%%%%%%%%%%%%%%%%%%%%%%%%%%%
%The $+$ and $/$ symbols are used to specify soft and hard priorities among tasks, respectively. 
%$^{\text{A}}\mathcal{T}_{\text{B}}$ denotes a Cartesian task of the frame $B$ relative to the frame $A$, while the $<<$ symbol denotes constraints.
%to the whole stack or to single levels of the stack.
%%%
Note that the first layer is responsible for the tracking of the planned lower-body trajectories.

\subsection{Angular Momentum Task}
An alternative stabilizing strategy consists in controlling the system's centroidal momentum \cite{orin2013centroidal} to maintain balance through emergent arm motions, thus without authoring any upper-body motion, as in Sec. \ref{subsec:raibert-like}. 
% \par
%Pre-multiplying both sides of the equation of the robot dynamics \eqref{eq:floatingdyn}, now including also the upper-body joints, by the term \mbox{${^{CoM}\bm{X}_{fb}}\mathbf{J}_{fb}^{-\top}$}, being \mbox{$\mathbf{J}_{fb} \in \mathbb{R}^{6 \times 6}$} the Jacobian of the virtual chain, and \mbox{${^{CoM}\bm{X}_{fb}}$} the adjoint matrix which change application point in a frame anchored at the Center of Mass (CoM), we obtain the  \textit{centroidal dynamics} relation \cite{orin2013centroidal}: 
%\begin{equation}
%    \label{eq:centroidal_dyn}
%    \bm{A}(\bm{q})\bm{\ddot{q}} + \bm{\dot{A}}(\bm{q},\bm{\dot{q}}) \bm{\dot{q}} = m \bm g + {\bm{G}_{\text{CD}}} \bm{F}_\text{C}
%\end{equation}
%where $\bm{A}(\bm{q}) \in\mathbb{R}^{6 \times (n+6)}$ is denoted as centroidal momentum matrix, which maps the velocities, and therefore momentum, of each individual body into the common reference frame at the CoM. The gravity acceleration is given by \mbox{$\bm g \in\mathbb{R}^{3}$}, while  \mbox{$m \in\mathbb{R}$} is the mass of the entire robot. Finally, \mbox{$\bm{G}_{\text{CD}}\in\mathbb{R}^{6\times k}$} is the centroidal dynamics grasp matrix.
%%%%%%%%%%%%%%%%%%%%%%%%%%%
% The left-hand side of \eqref{eq:centroidal_dyn} denotes
Let \mbox{$\bm{h} \in \mathbb{R}^6$} be the rate of change of the system centroidal momentum: 
%%%%%%%%%%%%%%%%%%%%%%%%%%%
\begin{equation}
    \bm{h} = \bm{A}(\bm{q})\bm{\dot{q}}
    \label{eq:centroidal_momentum_constraint}
\end{equation}
where $\bm{A}(\bm{q}) \in\mathbb{R}^{6 \times (n+6)}$ is the centroidal momentum matrix.
The centroidal momentum 
$\bm{h} = 
\begin{bmatrix}
 \bm{h}_{\text{lin}}^T & \bm{h}_{\text{ang}}^T
\end{bmatrix}^T$
is a spatial quantity comprised of the system's net linear momentum \mbox{$\bm{h}_{\text{lin}}\in \mathbb{R}^3$} and angular momentum \mbox{$\bm{h}_{\text{ang}}\in \mathbb{R}^3$} about the CoM.
%%%%%%%%%%%%%%%%%%%%%%%%%%%
%%%%%%%%%%%%%%%%%%%%%%%%%%%
\par
Similarly to \cite{wensing2013generation}, in order to maintain balance through emergent arm motions, 
the following angular momentum task $\mathcal{T}_{{\text{AngMom}}}$ can effectively provide a dampening of any excess angular momentum: 
\begin{equation}
    %\mathcal{T}_{{\text{AngMom}}} :\bm{h}_{\text{ang, d}}^{[P]} = 0
    \mathcal{T}_{{\text{AngMom}}} :\bm{h}_{\text{ang, d}} = 0
\end{equation}
%%%%%%%%%%%%%%%%%%%%%%%%%%
\par
The resulting SoT can be written as follows:
\begin{equation}
\small
\begin{pmatrix}
\left(\sum_i \   ^{\text{Pelvis}}\mathcal{T}_{{\text{Foot}}_i}^\text{[XYZ]}\right)/\\ 
%\\
%^{\text{World}}\mathcal{T}_{{\text{Pelvis}}}^\text{[P]}/\\ 
%\\
%\mathcal{T}_{{\text{AngMom}}}^\text{[P]} + \mathcal{T}_{\substack{\text{Posture}}}\\ 
^{\text{World}}\mathcal{T}_{{\text{Pelvis}}}/\\ 
%\\
\mathcal{T}_{{\text{AngMom}}} + \mathcal{T}_{\substack{\text{Posture}}}
\end{pmatrix}
<<
\begin{pmatrix}
\mathcal{C}_{\substack{\text{Pos.}\\\text{Lims}}}+\mathcal{C}_{\substack{\text{Vel.}\\\text{Lims}}}
\end{pmatrix},
\label{eq:box_stack_angular_momentum}
\end{equation}
where the task 
%$^{\text{World}}\mathcal{T}_{{\text{Pelvis}}}^\text{[P]}$ 
$^{\text{World}}\mathcal{T}_{{\text{Pelvis}}}$ 
is used to conveniently incorporate the IMU feedback as follows:
\begin{equation}
    % ^{\text{World}}\mathcal{T}_{{\text{Pelvis}}}^\text{[P]}: 
    %\omega_{\text{fb, d}}^{[P]} = -\omega_{\text{imu}}^{[P]}
    ^{\text{World}}\mathcal{T}_{{\text{Pelvis}}}: 
    \bm \omega_{\text{fb, d}} = - \bm \omega_{\text{imu}}
\end{equation}

\subsection{OCP Postural Task}
%%%%%%%%%%%%%%%%%%%%%%%%%%
Building upon the idea of treating the upper-body of a humanoid as a fly wheel in order to produce enough angular momentum to maintain balance, a dedicated OCP can be designed to produce a \textit{periodic} joint-space trajectory which maximizes the angular momentum along the tilting axis. 
%%%
The optimization is done on the fixed-based single-arm model of the robot, with 
\mbox{$\bm x = \begin{bmatrix}
 \bm q^T & \dot{\bm q}^T
\end{bmatrix}^T$}
and \mbox{$\bm u = \ddot{\bm q}$}. With abuse of notation here \mbox{$\bm q \in \mathbb{R}^{n_{\text{arm}}}$} is the single arm joint vector.
Some of the constraints in the optimization are:
\begin{itemize}
    \item[-] Same initial and final state, in order to produce a periodic trajectory. Note that, the the solver is let free to choose the arm's initial configuration.
    \item[-] Joint position, velocity and torque bounds. 
    \item[-] Work-space constraints on the end-effector Cartesian position, to prevent self-collisions.
\end{itemize}
The problem can be set up as DMS transcription with $N_s$ shooting intervals over a normalized time horizon, where the following cost function has been considered:
\begin{align}\label{eq:cost_function_posture_oc}
&F(\bm{X}, \bm{U}) =  
%%%
\sum_{\text{k}=0}^{N_s}
%\Big(\gamma_{\bm h_{\text{ang}}}(\bm h_{\text{ang,k}}^{[P]} - \bm h_{\text{d}}^{[P]}) + 
\Big(\gamma_{\bm h_{\text{ang}}}(\bm h_{\text{ang,k}} - \bm h_{\text{d}}) + 
 \gamma_{\bm{\dot q}}\bm{\dot q}_{\text{k}}^T\bm{\dot q}_{\text{k}}\Big)
\end{align}
in order to maximize the angular momentum e.g. along the \mbox{$y$-axis}.
%%%
Similarly to the fly wheel control in \cite{raibert1986legged}, the planned joint-space trajectory \mbox{$\bm q^*\in \mathbb{R}^{n_{\text{arm}}\times N_s}$} can be then replayed forward and backward on each robot arm with a velocity scaling proportional to the IMU feedback.
In order to do so, by considering an equivalent SoT to the one in \eqref{eq:box_stack_Raibert}, an upper-body postural task \mbox{$\mathcal{T}_{\substack{\text{Posture-OCP}}}$} is designed as follows:
\begin{equation}\label{eq:posture_oc}
\mathcal{T}_{\substack{\text{Posture-OCP}}}: \bm q_{\text{d}} = 
\bm q_{\text{k}}^*+ N_s(\tau_{\text{k+1}} - \text{k}/N_s)(\bm q_{\text{k+1}}^*-\bm q_{\text{k}}^*)
\end{equation}
where the scaling factor \mbox{$\tau_\text{k+1} \in [0, 1]$} is proportional to the IMU angular velocity controlled component, e.g. the pitch:
\begin{equation}
    \tau_{\text{k+1}} = \tau_{\text{k}} + \gamma_{\tau} \ \omega_{\text{imu}}^{[P]} 
\end{equation}
being the sample $\text{k} \in \mathbb{R}$ given by: 
$\text{k} =  \lfloor N_s \cdot \tau_{\text{k+1}} \rfloor$.

\section{Validation}\label{sec:simulations}
In order to first compare the performance of the proposed upper-body stabilizers, we set up a simulation benchmark scenario in Gazebo using the CENTAURO robot \cite{kashiri2019centauro} to perform the planned agile actions.
%%%%%%%%%%%%%%%%%%%%%%%%%%%%%%%%%%%%%%
CENTAURO is a $39$ DoF hybrid wheeled-legged quadruped equipped with a bimanual humanoid upper-body, and has a weight of  $92$ kg. The robot is fully torque-controlled with direct sensing of the link-side torque.
% Standard position control mode is also available. In its current setup no force/torque sensors are mounted at the robot end-effectors.
CENTAURO is powered by the \emph{XBotCore} middleware \cite{muratore2017xbotcore}, while the \emph{CartesI/O} framework \cite{laurenzi2019cartesi}, which relies on the hierarchical IK library \emph{OpenSoT} \cite{OpenSot17}, is responsible for Cartesian control.
%%%%%%%%%%%%%%%%%%%%%%%%%%%%%%%%%%%%%%%
Three separate ROS nodes running at \mbox{$1 \ kHz$} are responsible for the replay of the planned lower-body trajectories, the HIK managed by \emph{CartesI/O}, and the contact force distribution among the lower-body contact points, respectively. The joint-level controller runs at \mbox{$2 \ kHz$}.
%%%%%%%%%%%%%%%%%%%%%%%%%%%%%%%%%%%%%%%%%%%
%\par
Snapshots and time histories of the IMU pitch orientation are shown in Fig. \ref{fig:comparison}. 
These results are 
%summarized in \mbox{Table \ref{tab:comparison}} and 
illustrated in the provided supplementary video. 
Performing the squat jump without upper-body stabilization results in a notable tilt of the robot waist. Similarly, in the half-squat jump, the absence of upper-body stabilization results in an excessive tilt of the robot, eventually causing it to fall without accomplishing the task. In order to effectively mitigate the tilt rotation measured by \mbox{$\omega_{\text{imu}}^{[P]}$}, the shoulder pitch and elbow pitch joints have been enabled trough \mbox{$\bm{S}_{\text{ub}}$} in \mbox{$\mathcal{T}_{\substack{\text{Posture-Raibert}}}$}, see \eqref{eq:posture_raibert}, while the motion of the shoulder joints and the elbow joints is engaged in \mbox{$\mathcal{T}_{{\text{AngMom}}}$}. All the arm joints have been considered in \mbox{$\mathcal{T}_{\substack{\text{Posture-OCP}}}$}. 
%As it can be noticed from Fig. \ref{fig:jump_comparison} and Table \ref{tab:comparison}, performing the squat jump without upper-body stabilization results in a notable tilt of the robot waist, which can be effectively mitigated by the proposed stabilizers. Similarly, in the half-squat jump, the absence of upper-body stabilization results in an excessive tilt of the robot, eventually causing it to fall without accomplishing the task. 
%%%%%%%%%%%%%%%%%%%%%%%%%%%%%%%%%%%%%%%%%%%
% \par
Based on the benchmark results, %gathered in Table \ref{tab:comparison}
trading-off controller performance against number of parameters to tune and complexity of motion, i.e. number of enabled joints, the Raibert-like stabilizer stands out as the most suitable algorithm to perform experiments on the real robotic platform. In this respect, note that, although self-collisions are not inherently prevented, no collisions occurred in the performed agile actions for our choice of parameters. Due to high peak torques involved in the squat jump and battery current limitations, experimental validation has been 
% carried out and 
successfully performed on the half-squat jump task employing the Raibert-like stabilizing action, see Fig. \ref{fig:comparison_exp}.

\section{Conclusions} \label{sec:conclusions}
Aiming at performing agile actions with a centaur-type torque-controlled humanoid, this paper has presented a decoupled control architecture which meets the computational and implementation requirements to achieve a real demonstrator. Lower-body motion primitives are generated through optimal control in an offline stage, based on a simplified kinematic model. The planned lower-body trajectories are then replayed on the robot, in the online stage, using three different upper-body stabilizing strategies to maintain balance.
The stabilizers' performance has been compared in two types of simulated jumps, while experimental validation has been performed on a half-squat jump using CENTAURO. 
%%%%%%%%%%%%%%%%%%%%%%%%%%%%%%%%%%%%%%%
%\par 
%The planning problem of fast gaits represents a natural extension of this work. However, due to the wheeled-legged nature of CENTAURO, fast navigation on flat terrains can be already performed through wheeled motion. Alternative discretization algorithms, e.g. Direct Collocation, could be employed to speed up computations, although computation time is not a strict a requirement for the offline stage.
%% as well as experimental validation could be performed with the remaining proposed stabilizers.

%%%%%%%%%%%%%%%%%%%%%%%%%%%%%%%%%%%%%%%%%%%%%%%%%%%%%%%%%%%%%%%%%%%%%%%%%%%%%%%%
\bibliographystyle{IEEEtran}	
\bibliography{biblio}

\end{document}